\documentclass[pmlr]{jmlr}
\usepackage[T1]{fontenc}
\usepackage[utf8]{inputenc}

\usepackage{longtable}
\usepackage{booktabs}
\usepackage{siunitx}
\usepackage{bm}
\usepackage{csquotes}
\newcommand{\xhdr}[1]{{\noindent\bfseries #1}}

\usepackage{makecell}
\theorembodyfont{\upshape}
\theoremheaderfont{\scshape}
\theorempostheader{:}
\theoremsep{\newline}

\jmlrvolume{1}
\jmlryear{2022}
\jmlrworkshop{NeurIPS ML4CO Competition}

\newcommand{\tabincell}[2]{\begin{tabular}{@{}#1@{}}#2\end{tabular}}

\usepackage{lmodern}

\title[ML4CO Competition]{The Machine Learning for Combinatorial Optimization Competition (ML4CO): Results and Insights}

\author{
    \normalfont{Co-organizers}\\
    \Name{Maxime Gasse}\nametag{\thanks{Primary contact (maxime.gasse@polymtl.ca)}}, \Name{Simon Bowly}, \Name{Quentin Cappart}, 
    \Name{Jonas Charfreitag}, \\
    \Name{Laurent Charlin}, \Name{Didier Chételat}, \Name{Antonia Chmiela}, \Name{Justin Dumouchelle}, \\
    \Name{Ambros Gleixner}, \Name{Aleksandr M.\ Kazachkov}, \Name{Elias Khalil}, \Name{Pawel Lichocki}, \\
    \Name{Andrea Lodi}, \Name{Miles Lubin}, \Name{Chris J. Maddison}, \Name{Christopher Morris}, \\
    \Name{Dimitri J.\ Papageorgiou}, \Name{Augustin Parjadis}, \Name{Sebastian Pokutta}, \\
    \Name{Antoine Prouvost}, \Name{Lara Scavuzzo}, \Name{Giulia Zarpellon}\\
    ~\\
    \normalfont{Primal task winners}\\
    \Name{Linxin Yang}	\Email{yanglinxin@cuhk.edu.cn} \\
    \Name{Sha Lai}	\Email{221049039@link.cuhk.edu.cn} \\
    \Name{Akang Wang}	\Email{wangakang@sribd.cn} \\
    \Name{Xiaodong Luo} \Email{xiaodongluo@cuhk.edu.cn} \\
    \Name{Xiang Zhou} \Email{zhouxiang60@huawei.com} \\
    \Name{Haohan Huang} \Email{huanghaohan@huawei.com} \\
    \Name{Shengcheng Shao} \Email{shaoshengcheng@huawei.com} \\
    \Name{Yuanming Zhu} \Email{zhuyuanming5@huawei.com} \\
    \Name{Dong Zhang} \Email{zhangdong48@huawei.com} \\
    \Name{Tao Quan} \Email{quantao@huawei.com} \\
    ~\\
    \normalfont{Dual task winners} \\
    \Name{Zixuan Cao} \Email{caozixuan.percy@stu.pku.edu.cn} \\
    \Name{Yang Xu} \Email{1800010740@pku.edu.cn} \\
    \Name{Zhewei Huang} \Email{huangzhewei@megvii.com} \\
    \Name{Shuchang Zhou} \Email{zsc@megvii.com} \\
    ~\\
    \normalfont{Configuration task winners}\\
    \Name{Chen Binbin} 
    \Email{cbb18@mails.tsinghua.edu.cn} \\
    \Name{He Minggui} \Email{heminggui@huawei.com} \\
    \Name{Hao Hao} \Email{52194506007@stu.ecnu.edu.cn}\\
    \Name{Zhang Zhiyu} 
    \Email{zhangzhiyu6@huawei.com} \\
    \Name{An Zhiwu} 
    \Email{anzhiwu1@huawei.com} \\
    \Name{Mao Kun} 
    \Email{maokun@huawei.com} \\
}

\begin{document}

\maketitle
\newpage
\begin{abstract}
Combinatorial optimization is a well-established area in operations research and computer science. Until recently, its methods have focused on solving problem instances in isolation, ignoring that they often stem from related data distributions in practice. However, recent years have seen a surge of interest in using machine learning as a new approach for solving combinatorial problems, either directly as solvers or by enhancing exact solvers. Based on this context, the ML4CO aims at improving state-of-the-art combinatorial optimization solvers by replacing key heuristic components. The competition featured three challenging tasks: finding the best feasible solution, producing the tightest optimality certificate, and giving an appropriate solver configuration. 
Three realistic datasets were considered: \textit{balanced item placement}, \textit{workload apportionment}, and \textit{maritime inventory routing}. 
This last dataset was kept anonymous for the contestants.

\end{abstract}
\begin{keywords}
Combinatorial optimization, machine learning for combinatorial optimization
\end{keywords}

\section{Introduction}

The \textit{Machine Learning for Combinatorial Optimization} competition (ML4CO) aims at improving state-of-the-art combinatorial optimization solvers by replacing key heuristic components with machine learning models. The main scientific question is the following: \textit{``Is machine learning a viable option for improving traditional combinatorial optimization solvers on specific problem distributions, when historical data is available?''} 

While most combinatorial optimization solvers are presented as general-purpose, \textit{one-size-fits-all} algorithms, this competition focuses on the design of application-specific algorithms from historical data. This general problem captures a highly practical scenario relevant to many application areas, where a practitioner repeatedly solves problem instances from a specific distribution with similar patterns and characteristics. For example, managing a large-scale energy distribution network requires solving similar CO problems daily, with a fixed power grid structure while only the demand changes over time. This demand change is hard to capture by hand-engineered expert rules, and ML-enhanced approaches offer a possible solution to detect typical patterns in the demand history. Other examples include crew scheduling problems that have to be solved daily or weekly with minor variations or vehicle routing where the traffic conditions change over time, but the overall transportation network does not. 

The competition features three challenges for machine learning.
Each of them corresponds to a specific control task arising in the open-source solver \textit{SCIP}~\citep{GamrathEtal2020OO} and is exposed through a unified OpenAI Gym-like API based on the Python library \textit{Ecole}~\citep{prouvost2020ecole}. The three challenges are as follows: (1) \textit{a primal task}, consisting of producing the best feasible solution, (2) \textit{a dual task}, consisting of producing the best branching decisions, and (3) \textit{a configuration task}, consisting of finding the best parameters before calling the solver. For each challenge, participants were evaluated on three problem benchmarks originating from diverse application areas, each represented as a collection of \textit{mixed-integer linear program} (MILP) instances. 


\section{Datasets}

The participants' solutions were evaluated on three problem benchmarks from diverse application areas for each challenge. Each participant submitted a decision-making code, i.e., an algorithmic solution or trained ML model,  
for each of the benchmarks, or a single code that works for all benchmarks. A problem benchmark consists of a collection of MILP instances in the standard MPS file format. Each benchmark was split into a training and a test set, originating from the same problem distribution. While the training instances were made public at the beginning of the competition for participants to train their models,  the test instances were kept hidden for evaluation purposes and were only revealed at the end of the competition. 
The first two problem benchmarks were inspired by real-life applications of large-scale systems at Google, while the third benchmark was presented to the participants as an anonymous problem inspired by a real-world, large-scale industrial application. The dataset used are publicy available on Github.\footnote{\url{https://github.com/ds4dm/ml4co-competition}}

\xhdr{Benchmark 1: Balanced Item Placement} This problem involves spreading items, e.g., files or processes, across containers, e.g., disks or machines, utilizing them evenly. Items can have multiple copies, but at most, one copy can be placed in a single bin. The number of items that can be moved is constrained, modeling the real-life situation of a live system for which some placement already exists. Each problem instance is modeled as a MILP, using a multi-dimensional, multi-knapsack formulation. This dataset contains 10000 training instances (pre-split into 9900 train and 100 validation instances). 

\xhdr{Benchmark 2: Workload Apportionment} This problem deals with apportioning workloads, e.g., data streams, across as few workers, e.g., servers, as possible. The apportionment is required to be robust to any worker's failure. Each instance problem is modeled as a MILP, using a bin-packing with an apportionment formulation. This dataset contains 10000 training instances (pre-split into 9900 train and 100 validation instances). 

\xhdr{Benchmark 3: Maritime Inventory Routing (Anonymous Problem)} This problem plays an integral role in global bulk shipping. The instances corresponding to this benchmark are assembled from a public dataset \citep{papageorgiou2014mirplib}, whose origin was kept secret to prevent cheating. Reverse-engineering for the purpose of recovering the test set was  forbidden. The dataset contains 118 training instances (pre-split into 98 train and 20 validation instances).

\section{Evaluation Metrics}

Each of the three challenges is associated with a specific evaluation metric reflecting a different objective. We describe how each metric is computed over a single problem instance. The final goal of the participants is to optimize this metric in expectation over a hidden collection of test instances.
Because the evaluation metrics are time-dependent, 
all evaluations were run on the same hardware setup, using an Intel Xeon processor with 2.4GHz, 20GB of RAM, and an Nvidia Tesla  V100-8G GPU with 8GB of GPU memory. A maximum time budget was given for each task to process each test instance (5, 15, and 15 minutes for the primal, dual, and configuration tasks, respectively),  after which the environment was terminated. By doing so, participants were asked to focus on making both good and fast decisions. 

In general, a MILP instance is expressed as follows,

\begin{eqnarray}
\underset{\mathbf{x}}{\operatorname{arg\,min}} \quad \mathbf{c}^\top\mathbf{x} &&  \notag \\
\text{subject to} \quad \mathbf{A}^\top\mathbf{x} & \leq & \mathbf{b}  \notag \\
\mathbf{x} & \in & \mathbb{Z}^p \times \mathbb{R}^{n-p}, \notag
\end{eqnarray}
where $\mathbf{c} \in \mathbb{R}^n$ denotes the coefficients of the linear objective, $\mathbf{A} \in \mathbb{R}^{m \times n}$ and $\mathbf{b} \in \mathbb{R}^m$, denote the coefficients and upper bounds of the linear constraints, respectively, while $n$ is the total number of variables, $p \leq n$ is the number of integer-constrained variables, and $m$ is the number of linear constraints. We used the following three metrics for the evaluation.

\xhdr{Primal Integral} 
This metric measures the area under the curve of the solver's primal bound, i.e., its global upper bound, which corresponds to the value of the best feasible solution found so far. 
By providing better feasible solutions over time, the value of the primal bound decreases. 
With a time limit $T$, the primal integral is expressed as follows,
\begin{equation*}
\int_{t=0}^{T} \mathbf{c}^\top \mathbf{x}^\star_t\;\mathrm{d}t - T\mathbf{c}^\top\mathbf{x}^\star,
\end{equation*}
where $\mathbf{x}^\star_t$ is the best feasible solution found at time $t$, so that $\mathbf{c}^\top \mathbf{x}^\star_t$ is the primal bound at time $t$, and $T\mathbf{c}^\top\mathbf{x}^\star$
is an instance-specific constant that depends on the optimal solution $\mathbf{x}^\star$.
The primal integral is to be \textit{minimized}, 
and takes an optimal value of 0.
To compute this metric unambiguously, a trivial initial solution $x^\star_0$ is always provided to the solver at the beginning of the solving process. 
Also, the constant term $\mathbf{c}^\top\mathbf{x}^\star$ can be safely ignored at training time when
participants train their control policy. However, when we evaluated the
participant submissions at test time, this constant term, or a proper substitute, was incorporated in the reported metric.

\xhdr{Dual Integral} 
This metric measures the area over the curve of the solver's dual bound, i.e., its  global lower bound, which usually corresponds to a solution of a valid relaxation of the MILP. 
By branching, the linear relaxation corresponding to the branch-and-bound tree's leaves gets tightened, and the dual bound increases over time. 
With a time limit $T$, the dual integral is expressed as follows,
\begin{equation*}
T\mathbf{c}^\top\mathbf{x}^\star - \int_{t=0}^{T} \mathbf{z}^\star_t\;\mathrm{d}t,
\end{equation*}
where $\mathbf{z}^\star_t$ is the best dual bound at time $t$, and $T\mathbf{c}^\top\mathbf{x}^\star$ is
an instance-specific constant that depends on the optimal solution value $\mathbf{c}^\top\mathbf{x}^\star$.
The dual integral is to be \textit{minimized}, and takes an optimal value of 0. 

In the context of branching, this metric is unambiguous to compute, as the root node of the tree always provides an
initial dual bound $\mathbf{z}^*_0$ at the beginning of branching.  The constant term $\mathbf{c}^\top\mathbf{x}^\star$ can be safely ignored for training, but it was incorporated in the evaluation metric.

\xhdr{Primal-Dual Gap Integral}
This metric measures the area between two curves, 
the solver's \textit{primal bound} and \textit{dual bound}. 
Hence, the metric benefits both from improvements obtained from the primal side
(finding good feasible solutions), and on the dual side (producing a tight optimality certificate).
With a time limit $T$, 
the primal-dual gap integral is expressed as follows,
\begin{equation*}
\int_{t=0}^{T} \mathbf{c}^\top \mathbf{x}^\star_t - \mathbf{z}^\star_t\;\mathrm{d}t.
\end{equation*}
The primal-dual gap integral is to be \textit{minimized}, and takes an optimal value of 0.
In the context of algorithm 
configuration, an initial value is required for the two curves at time $t=0$. 
Therefore, an initial trivial solution $x^\star_0$ and a valid initial
dual bound $z^\star_0$ are always provided to the solver for this task.
 

\section{Solving the Primal Task (Winning Solution by CUHKSZ\_ATD)}

The primal task deals with finding good primal solutions at the root node of the branch-and-bound tree \citep{khalil2017learning,nazari2018reinforcement,li2018combinatorial,nair2020solving}. To that end, the environment (SCIP solver) does not perform any branching but enters an infinite loop at the root node, 
which collects the solutions proposed by the agents, 
evaluates their feasibility and updates the overall best solution reached so far, thus lowering the current primal bound (upper bound). 
The metric of interest for this task is the primal integral, which considers the rate at which the primal bound decreases over time. 
To model a realistic scenario, each problem instance has been preprocessed by SCIP (problem reduction, cutting planes, etc.).
Moreover, the root linear program (LP) relaxation was solved before the participants were asked to produce feasible solutions. 
To prevent SCIP from searching for primal solutions by itself, all primal heuristics were deactivated. 
Further, to compute this metric unambiguously, even when no solution has been found yet, an initial primal bound (trivial solution value) was provided for each instance,  which is to be given to SCIP at the beginning of the solving process in the form of a user objective limit. Execution time was limited to five minutes. We exploit the problem structures and tackle item placement, workload apportionment, and anonymous problems by utilizing classic primal heuristics in a more judicious manner.

\subsection{Balanced Item Placement} 
Let $I$ denote the set of items and $J$ denote the set of containers. 
Let a binary variable $x_{ij}$ be $1$ if item $i$ is placed in container $j$ and $0$ otherwise.
Each item will be placed in exactly a single container, as shown by constraints~(\ref{eq:ip:assignment}). 
Let $K$ represent the set of dimensions.
For dimension $k \in K$ of container $j \in J$, knapsack constraints~(\ref{eq:ip:knapsack}) represent some physical considerations while (\ref{eq:ip:define_y}) and~(\ref{eq:ip:define_z}) properly account for the placement unevenness, which is penalized in the objective~(\ref{eq:ip:obj}). 
\vspace{-10pt} 
\begin{align}
	& \underset{x, y, z}{\text{min}} && \sum_{j \in J}^{} \sum_{k \in K}^{} \alpha_{k} y_{jk} + \sum_{k \in K}^{} \beta_k z_{k} \label{eq:ip:obj} \\   \noalign{\vskip-4pt}
	& \; \text{s.t.}  && \sum_{j \in J}^{} x_{ij} = 1 &&& \forall i \in I \label{eq:ip:assignment} \\  \noalign{\vskip-4pt}
	&              && \sum_{i \in I }^{} a_{ik} x_{ij} \leq b_{k}    &&& \forall j \in J, \forall k \in K  \label{eq:ip:knapsack}  \\  \noalign{\vskip-4pt}
	&              && \sum_{i \in I}^{} d_{ik} x_{ij} + y_{jk}  \geq 1   &&& \forall j \in J, \forall k \in K \label{eq:ip:define_y} \\  \noalign{\vskip-4pt}
	&              && y_{jk} \leq z_{k}    &&& \forall j \in J, \forall k \in K  \label{eq:ip:define_z} \\   \noalign{\vskip-4pt}
	&              && x_{ij} \in \left\{0, 1\right\} &&& \forall i \in I, \forall j \in J \\   \noalign{\vskip-4pt}
	&              && y_{jk} \geq 0         &&& \forall j \in J, \forall k \in K  \label{eq:ip:positive}
\end{align}

We analyze $10, 000$ item placement instances and find out that: (i)~$|I| = 105, |J| = 10$; (ii)~$a_{ik}, d_{ik}$ values of $5$ items are big and placing any two of them in the same container would incur large penalty. 
Based on this empirical finding, we place the $5$ big items into the first five containers respectively before applying any primal heuristics. 

\xhdr{Meta-heuristics}
We apply a greedy method in which items are first sorted based on their sizes and then assigned to containers.
This will produce the very first feasible solution.
After that, we select one or two items respectively from two containers and swap them if this leads to a better incumbent.  

\xhdr{Math-heuristics} 
We consider a construction method and an improvement method, based on solving mathematical models. 
The construction method consists of two steps:
(i)~first assign items to the first five containers by solving an assignment model; (ii)~then assign the remaining items to the last five containers by solving a sub-MIP.
For solution improvement, we properly choose two out of the last five containers, and then solve a sub-MIP to reassign items within those two containers optimally.

\subsection{Workload Apportionment}
Let $M$ and $N$ denote the set of tasks and the set of machines, respectively.
For task $i \in M$, only a subset of machines, denoted by $N^i \subseteq N$, are accessible. 
Let a binary variable $y_j$ be $1$ if machine $j$ is used and $0$ otherwise.
Let $x_{ij}$ denote the amount of workload from task $i$ to machine $j$, as defined in constraints~(\ref{eq:lb:define_x}).
Constraints~(\ref{eq:lb:capacity}) enforce the capacity requirement for each machine.
The apportionment is required to be robust to any one machine's failure, as indicated by constraints~(\ref{eq:lb:robust}). 
\begin{align}
	& \underset{x, y}{\text{min}} && \sum_{j \in N}^{} y_j  \label{eq:lb:obj} \\   \noalign{\vskip-4pt}
	& \; \text{s.t.}  &&  x_{ij} \leq a_i y_j &&& \forall i \in M, \forall j \in N^i  \label{eq:lb:define_x} \\ \noalign{\vskip-4pt}
	&              && \sum_{i \in M: j \in N^i}^{} x_{ij} \leq b_j    &&& \forall j \in N   \label{eq:lb:capacity}  \\ \noalign{\vskip-4pt}
	&              && \sum_{j \in N^i \setminus \left\{j^\prime\right\}}^{}  x_{ij} \geq a_i   &&& \forall i \in M, \forall j^\prime \in N^i  \label{eq:lb:robust}  \\ \noalign{\vskip-4pt}
	&              && y_j \in \left\{0, 1\right\}   &&& \forall j \in N \\ \noalign{\vskip-4pt}
	&             && 0 \leq x_{ij} \leq b_{j}  &&& \forall i \in M, \forall j \in N^i  \label{eq:lb:positive}  
\end{align} 
Rounding up a solution to the linear programming relaxation of model (\ref{eq:lb:obj})~--~(\ref{eq:lb:positive}) would produce a feasible solution to (\ref{eq:lb:obj})~--~(\ref{eq:lb:positive}).
To further exploit the possibility of rounding a fractional solution towards a new incumbent, we choose a rounding threshold parameter $\eta$ in an adaptive manner and round up $y_j$ only if it exceeds $\eta$. 
Specifically, we first select a target objective value based on the current primal and dual bounds and then determine $\eta$ via quantile selection such that after rounding with $\eta$ the objective matches the pre-determined value. 
If the rounding step produces a new incumbent solution, we then update the primal bound; otherwise, we set the dual bound to the corresponding objective value. 
We iterate this process until the primal-dual gap falls below a predetermined value. In model~(\ref{eq:lb:obj})~--~(\ref{eq:lb:positive}), constraints~(\ref{eq:lb:capacity}) can be tightened as follows:
\vspace{-10pt}
	\begin{equation}
		\begin{aligned}
			\sum_{i \in M: j \in N^i}^{} x_{ij} \leq b_j y_j    &&& \forall j \in N.  
		\end{aligned}  \label{eq:lb:tightened_capacity}
		\vspace{-10pt}
	\end{equation}
Furthermore, across $10, 000$ instances, we observe that $a_i < b_j$ for $i \in M, j \in N^i$. 
As a result, constraints~(\ref{eq:lb:define_x}) are dominated by~(\ref{eq:lb:tightened_capacity}) and thus eliminated from the model.
Now we call~(\ref{eq:lb:obj}), (\ref{eq:lb:robust})~--~(\ref{eq:lb:tightened_capacity}) as the \enquote{tightened model}. 

In our implementation, we first apply the rounding heuristic method to the root LP solution to the original model and then to the optimal LP solution to tightened model.
We then use RINS~\citep{danna2005exploring} to further improve the incumbent. 
In particular, we define and solve a sub-MIP based on the tightened model, using its LP solution and the current incumbent as a guide to fix part of the binary variables. 

\subsection{Anonymous Problem}
Though the concrete MILP formulation is not available, we discover some pattern from anonymous instances in the LP file format.
In particular, one can define a planning horizon and associate with each discrete variable a time period.
The details can be deduced from the constraint hypergraph~\citep{rossi2006handbook} in which every node represents a discrete variable and every edge joins a pair of variables if they occur together in a constraint.
Let $H$ denote the planning horizon and $h \in H$ denote a time period. We use a heuristic called the \enquote{rolling-horizon} method to generate our final high quality primal solution. It consists of the following steps: (i)~ignore constraints involving discrete variables with their $h$ values greater than $\widetilde{H}$; (ii)~relax the integrality constraints on variables with their $h$ values greater than $\overline{H}$; (iii)~fix those discrete variables with their $h$ values less than $\widehat{H}$ at the optimal solution from a previous run; (iv)~solve the sub-MIP;(v)~increase $\widehat{H}, \overline{H}$ and $\widetilde{H}$ adaptively and then iterate steps (i) - (iv) until $\widehat{H} = H$.  Before calling the computationally expensive rolling-horizon method, we utilize feasibility pump~\citep{fischetti2005feasibility, bertacco2007feasibility} to generate the very first solutions and call RENS~\citep{berthold2014rens} once to improve that solution.
The RENS model is a sub-MIP defined by fixing those discrete variables with their $h$ values less than $0.9H$ at the incumbent.

\section{Solving the Dual Task (Winning Solution by Nuri)}

The dual task deals with obtaining tight optimality guarantees (dual bounds) with branching \citep{khalil2016learning,balcan2018learning,conf/nips/GasseCFCL19,gupta2020hybrid,cappart2021combining}. 
Making good branching decisions is regarded as a critical component of modern branch-and-bound solvers.
However, it has received little theoretical understanding to this day \citep{lodi2017learning}. 
In this task, the environment runs a full-fledged branch-and-cut algorithm with SCIP, and the participants only control the branching decisions of the solver. 
The metric of interest is the dual integral, which considers the rate at which which the dual bound increases over time. Fruther, all primal heuristics are deactivated to focus only on proving optimality using branching. Execution time was limited to 15 minutes.

We propose Knowledge Inheriting Dataset Aggregation (KIDA), which trains a neural model to decide branching variables with an enhanced version of Dataset Aggregation and a surprisingly effective Model Weight Averaging (MWA) trick. KIDA consists of three steps. First, candidate models are trained by combining the ideas of DAgger~\citep{ross2011reduction} and Born-Again Neural Networks~\citep{furlanello2018born} to imitate the Strong Branching~\citep{achterberg2005branching} heuristics on an expanded dataset. Then, we apply the MWA trick to derive more candidate models. Finally, the model with the highest cumulated reward on the validation set is selected as the final model to be used for testing in the deployment environment. We demonstrate that KIDA achieves top performance on the benchmarks of Balanced Item Placement and Anonymous Problem with a single model, surpassing methods that purely imitate Strong Branching heuristics.

\xhdr{Limitations of Imitating Strong Branching} Classical approaches~\citep{conf/nips/GasseCFCL19, nair2020solving} to train neural models to decide branching variables rely on imitation of Strong Branching heuristics. However, on the two benchmarks of Balanced Item Placement and Anonymous Problem, we find that having good performance measured by the accuracy of imitating Strong Branching (SBA) does not always lead to good performance on the deployment environment measured as cumulated reward (CR), as shown in Table~\ref{table:epochs}. On the other hand, though CR is a reliable evaluation metric that is highly consistent between validation and deployment, as shown in Table~\ref{table:epochs}, it is a sparse signal and cannot be used directly in an Imitation Learning framework. In light of these, we propose to first train models to imitate Strong Branch heuristics, and then use a greedy search to select the final model from trained models and their derived weight averaging models.
We also note on the benchmark of Workload Apportionment, SBA almost completely fails to correlate with the deployment-time CR as models with high SBA are often inferior to simple random strategy, as shown in Table~\ref{table:cum_reward}.

\begin{table}[!htbp]
\centering  
\caption{{Performance Comparison of Models of Different Epochs on the benchmark of Balanced Item Placement}. SBA is not strictly related to CR}
\label{table:epochs}
\resizebox{.55\textwidth}{!}{
\begin{tabular}{ccccc}
\toprule
Epoch    & Top 1 SBA     & Top 3 SBA   & \tabincell{c}{Cum. Reward on \\ Validation Set} & \tabincell{c}{Cum. Reward on \\ Test Set}     \\ \midrule
1 & 0.780 & 0.932            & 5202.6 & 5028.6          \\
5 & 0.803          & \textbf{0.948}                    & \textbf{5545.7}     & \textbf{5289.3}\\ 
10 & 0.808          &  0.946              & 5131.9 & 4807.9         \\ 
20 & \textbf{0.810}         & 0.947              & 5038.8 & 4840.7         \\ \bottomrule
\end{tabular}}
\end{table}

\xhdr{Enhanced Dataset Aggregation} We find that Dataset Aggregation (DAgger)~\citep{ross2011reduction} improves SBA of trained models and helps to get better final performance in the deployment environment. We are also inspired by Born-Again Neural Networks~\citep{furlanello2018born} to utilize all trained models from the whole training process. The training of the current model depends on the data generated with the last model. By the end of this step, we have a collection of candidate models that can be leveraged in the next step.

\xhdr{Model Weight Averaging and Greedy Search} In the setting of Dual Task, Model Ensembles~\citep{dietterich2000ensemble, kidzinski2018learning}, which simply averages outputs of multiple models, doesn't work because the increase in run time by using multiple models will lead to fewer rounds of interaction with the underlying SCIP solver, and consequently degrading performance. In our experiments, averaging the weights of models trained in different epochs~\citep{tarvainen2017mean} doesn't have any positive effect either. In contrast, we consider building a new model $\pi_{avg}$ by averaging the weights of different trained models during the DAgger process. Formally, for $\Omega$ models obtained from DAgger with parameters $(\theta_{0}, \theta_{1}, ..., \theta_{\Omega-1})$, the parameters of $\pi_{avg}$ are obtained by:
$\theta_{avg}=\sum_{i=0}^{\Omega-1}\theta_{i}/\Omega$. In practice, we perform a greedy grid search over different model combinations and different $\Omega$ to select the best averaged model with the highest CR. Note original models are also included in the search as the particular case of $\Omega=1$.

\xhdr{Results}

The final performance of our methods on benchmarks is shown in Table \ref{table:cum_reward}. Baseline results are strictly from the official code. For KIDA, we select the best-performing model by greedy search over $\Omega={1,2,...,5}$.

\begin{table*}[htbp]
\centering
\caption{The Cumulated Reward in Three Benchmarks of Different Methods}

\label{table:cum_reward}
\vspace{2.5mm}
\resizebox{.55\textwidth}{!}{
\begin{tabular}{lllll}
\toprule
\multicolumn{1}{c}{}                  & Random & \multicolumn{1}{c}{Baseline} &  KIDA   \\ \midrule
Balanced Item Placement               & 3300.7      & 4937.8                                                      & \textbf{7561.6}~($\bm{\Omega=3}$)        \\
Workload Apportionment                & \textbf{624928.9}    & 624043.6                                                      & 623996.0     \\
\multicolumn{1}{l}{Anonymous Problem} & 31145708.4      & \multicolumn{1}{c}{30965031.6} & \textbf{32832618.7}~($\bm{\Omega=1}$) \\ \bottomrule
\end{tabular}}
\end{table*}

\noindent\textbf{Balanced Item Placement}: a KIDA model with $\Omega=3$ overtakes the other methods with a significant margin. 

\noindent\textbf{Workload Apportionment}: surprisingly, the random strategy is significantly better than the baseline model in this benchmark. This may be attributed to the failure of Strong Branching heuristics for this set. 

\noindent\textbf{Anonymous Problem}: the best KIDA model we find is a single DAgger model without averaging. Applying KIDA with $\Omega \geq 2$ leads to a slight decline in deployment-time CR. 

\section{Solving the  Configuration Task (Winning Solution by EI-OROAS)} 
The configuration task deals with deciding on a good parameterization of the solver for a given problem instance \citep{hutter2011sequential}. 
The environment required for this task is more straightforward than for the two previous ones since it involves only a single decision for the agents, i.e., contextual bandit problem). Participants were allowed to tune any of the existing parameters of SCIP. They could choose between providing a fixed set of parameters that work well on average for each problem benchmark or producing instance-specific parameterizations based on the characteristics of each instance. The metric of interest for this task was the primal-dual gap integral, which combines both improvements from the dual and from the primal side over time. To compute this metric unambiguously, even when no primal or dual bound exists, an initial primal bound (trivial solution value) and an initial dual bound (pre-computed root LP solution value) for each instance are both provided. Execution time was limited to 15 minutes.

The configuration task deals with deciding on a good parameterization of the solver for a given problem instance. Specifically, there are two questions need to answer: 1. How to search good parameters for seen instances? 2. How to recommend parameters for unseen instances? With a basic exploration, we conclude the following challenges: 1) Numerous, heterogeneous and conditional dependent configurable parameters 2) Expensive optimization cost for open instances and Limited samples for anonymous set. In our solution, we proposed a novel space reduction pipeline and adopted HEBO~\citep{cowen2020empirical} as optimizer to search good parameter, while a ML-based classifier to recommend parameters. 

\xhdr{Implementation of method} First, the search space is reduced to accommodate Bayesian optimization. 
Expert experiences were introduced to filter insignificant parameters. Then, fitting a XGboost observer to tighten the region of candidate parameters again with Gini coefficients. 
The search space $\Omega$ will be split to $k$ sub-space (om)at the start, then optimized one by one with broad first search order. Algorithm~\ref{alg:config_task_framework} presents the pseudo-code for search. Specifically, a two-layer loop is used to solve the problem. The outer loop optimizes each sub-space, and the inner loop is a standard BO method that minimizes the objective function. Some comments are all as follows. \textit{1) Full solution initialization:} The $x^*$ is assigned the SCIP default value for initialization and replaced by optimal partial solution $x^i~(i=1,\cdots,k)$ until the end of the iterations. \textit{2) Partial solution initialization:} The sub-space $\varOmega_i$ is defined as the current search space, and the initial dataset $\mathcal{D}_0^{i}$ is sampled in this space. 
\textit{3) Model construction and optimization:} The surrogate model construction and acquisition function optimization is following the HEBO. \textit{4) Objective evaluation:} The new suggested solutions $x_{1:q}^{i}$ will supplemented by $x^{*}$ into full solutions, configuring for SCIP, and solving for all instances in $\mathcal{I}$. the mean value of the Primal-dual gap integral for all instances is returned as an objective value. \textit{5) Update full solution:} When the inner loop ends, the optimal solutions $x_i$ will update the global full optimal solution $x^*$.
\begin{table}[htbp!]
  \caption{The statistical results of primal-dual gap by default and tuned parameters of SCIP.}

\centering
\resizebox{.55\textwidth}{!}{
\begin{tabular}{cccc}
  \toprule
   Dataset & \makecell[c]{Default \\Performance} & \makecell[c]{Tuned \\ Performance} &\makecell[c]{Improvement} \\ \midrule
   Item Placement & 16942 & \textbf{8781} & $\times$1.92\\ 
   Load Balancing & 22168 & \textbf{9499} & $\times$2.33\\ 
   Anonymous-c1 & 277e+4 & \textbf{257e+4} & $\times$1.08\\ 
   Anonymous-c2 & 1133e+6 & \textbf{730e+6} & $\times$1.55 \\ 
   Anonymous-c3 & 726e+4 & \textbf{574e+4} & $\times$1.26\\ 
  \bottomrule
  \end{tabular}}
  \label{fig:table_1}
\end{table}
\begin{algorithm2e}[htbp]
  \scriptsize 
  \SetKwInOut{Input}{Input}\SetKwInOut{Output}{Output}
  \caption{Pseudocode of Parameter Adaptive Bayesian Optimization}\label{alg:config_task_framework}
  \Input{$\Omega=\{ \varOmega_1,\varOmega_2,\cdots,\varOmega_k \}$~: Search space;
  $\mathcal{I}$ : Instances set;
  $N$ : Total number of iterations of each sub-search space
  }
  $x^* \leftarrow$ SCIP default value in $\Omega$. \tcc*[f]{full solution initialization}\\
  \For{$i \leftarrow 1~\KwTo~k$}{
    Initialize $\mathcal{D}_0^{i}$ by random sample from sub-search space $\varOmega_i$. \tcc*[f]{partial solution initialization}\\
    \For{$j \leftarrow 0~\KwTo~N-1$}{
      Fit a surrogate model to current dataset $\mathcal{D}^{i}_{j}$. \tcc*[f]{modeling}\\ 
      Find $q$ solutions $x_{1:q}^{i}$ by maximizing three acquisition functions. \tcc*[f]{suggest}\\
      Evaluate new partial solutions ($x^i_{1:q}$) by querying the  function to get $y_{1:q}=f(x_{1:q}^{i},x^{*},\mathcal{I})$. \tcc*[f]{evaluate}\\
      Update the dataset creating $\mathcal{D}^{i}_{j+1}=\mathcal{D}^{i}_{j}\cup \{ x_l^{i},y_l\}_{l=1}^{q}$. \tcc*[f]{tell}\\
    }
    $x^* \leftarrow x^*\cup argmin_{x^i\in \mathcal{D}^i}.$\tcc*[f]{update full optimum solution}
  }
  \Output{$x^*$~(the best-performing parameters set for SCIP solver on $\mathcal{I}$).}
  \end{algorithm2e}

\xhdr{Empirical study} We solve the problem as described above, where the anonymous dataset is divided into three clusters according to the features. The statistical results of SCIP solver with default and tuned parameters are summarize in Table~\ref{fig:table_1}. The performance of the SCIP has been improved, with a maximum of 2.33 times improvement on Load Balancing dataset and minimum of 1.08 times on Anonymous-c1. The experimental results demonstrate the effectiveness and robustness of our method in the MIP solver configuration.

\section{Conclusion}
The goal of this competition was to foster the design of innovative methods to improve state-of-the-art combinatorial optimization solvers by replacing key heuristic components with machine learning models. 
To that, we proposed three challenging benchmarks. In total, we received 12 for the primal task 12, 23 for the dual task, and 15 for the configuration task. We provide descriptions of some of the participants' solutions on the competition website.\footnote{\url{https://www.ecole.ai/2021/ml4co-competition/}}
The results indicate that machine learning for combinatorial optimization has potential, although more work must be done before it becomes relevant for practical, real-world use. We plan to maintain this competition across the years to monitor performance improvements over the years.

\vspace{-2pt}
\acks{The event was sponsored by the Artificial Intelligence Journal, as well as Compute Canada, Calcul Québec, and Westgrid who graciously provided the compute resources and the prize money. We finally thank all the participants.}

\bibliography{ml4co}

\appendix





\end{document}